\documentclass{article}
\usepackage{arxiv}           
\usepackage[numbers]{natbib} 

\newcommand{\mat}[1]{\mathbf{#1}}  
\def\Q{\mat{Q}_i}
\def\K{\mat{K}_i}
\def\V{\mat{V}_i}
\def\P{\mat{P}_i}
\def\O{\mat{O}_{i-1}}
\def\M{\mat{M}_i}
\def\u{\vec{u}}
\def\x{\vec{x}}
\def\y{\vec{y}}
\def\z{\vec{z}}

\title{KV-weights are all you need for skipless transformers}

\author{Nils Graef\thanks{\texttt{info@openmachine.ai}} \\
  \href{https://openmachine.ai}{OpenMachine}}

\begin{document} \maketitle

\begin{abstract}
\citet{simplified} detailed a skipless transformer without the V and P (post-attention projection) linear layers, which reduces the total number of weights. However, this scheme is only applicable to MHA (multi-head attention) \cite{vanilla}, but not for MQA (multi-query attention) \cite{MQA} and GQA (grouped-query attention) \cite{GQA}. The latter schemes are used by many popular LLMs such as Llama 2, Mistral, Mixtral, PaLM, and Gemma \cite{Llama2, mistral, mixtral, PaLM, gemma}. Therefore, this micro-paper \cite{micro-paper} proposes mathematically equivalent versions that are suitable for MQA and GQA. For example, removing Q and P from a skipless version of Mistral-7B would remove 15\% of its weights, and thus reduce its compute and memory complexity. Watch our explainer video \citep{remove-video} and see \citep{tricks, precompute} for code and more transformer tricks.
\end{abstract}

\section{Vanilla transformer without skip connections}

\begin{figure}[h!] \centering 
  \includegraphics[scale=0.87]{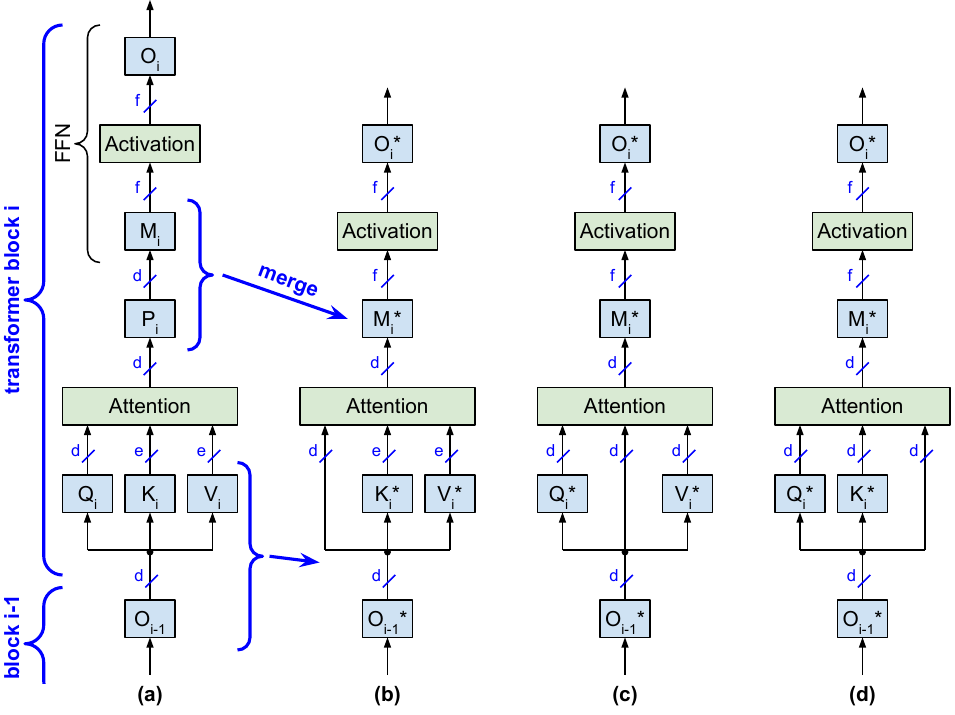}
  \caption{(a) Skipless vanilla transformer; equivalent versions with (b) Q and P merged into the FFN (feedforward network); (c) K and P merged into FFN; (d) V and P merged into FFN. $\M^*, \Q^*, \K^*, \V^*, \O^*$ are defined in table \ref{tab1}.}
\label{fig1} \end{figure}

\citet{skipless} have shown how transformers without skip connections and normalization (Figure \ref{fig1}(a)) can be trained successfully. Removing skip connections and normalization allows us to merge linear layers in a mathematically identical way as shown in Figures \ref{fig1}(b) to (d). This reduces the number of weights without changing the functionality as follows:
\begin{itemize}[topsep=-1pt, itemsep=-1pt]
  \item Figure \ref{fig1}(b) is mathematically identical to Figure \ref{fig1}(a) and eliminates $2d^2$ weights per transformer block by merging $\P$ into $\M^*$ and $\Q$ into $\O^*$.
  \item For MHA where $e = d$, Figures \ref{fig1}(c) and (d) are mathematically identical to Figure \ref{fig1}(a) and eliminate $2d^2$ weights per transformer block by merging $\P$ into $\M^*$ and $\K$ or $\V$ into $\O^*$.
  \item This requires that $\Q, \K, \V$ are invertible (i.e. nonsingular). It is extremely rare that a square matrix with random values is not invertible \cite{invertible} (which requires its determinant to be exactly 0).
\end{itemize}

Figure \ref{fig1} uses the following dimensions and weight matrices, based on the type of attention:
\begin{itemize}[topsep=-1pt, itemsep=-1pt]
  \item $d$: embedding dimension
  \item $e$: $e = d$ for MHA. For MQA, $e = d / n_{heads}$. And for GQA, $e = d \cdot n_{kv\_heads} / n_{heads}$.
  \item $f$: hidden dimension of the FFN. $f = 4d$ in the vanilla transformer; \citet{MQA} uses $f > 4d$. For models that use a GLU variant \cite{GLU} (such as Llama and Mistral), the effective $f'$ for the first linear layer M is $f' = 2f$, because the GLU variant uses two linear layers that are combined (via pointwise multiplication) with a non-linear activation function.
  \item $\Q, \K, \V, \P$: The weight matrices of the linear layers for query, keys, values, and the post-attention projection of transformer block $i$.
  \item $\M, \mat{O}_i$: The weight matrices of the FFN input and output linear layers.
\end{itemize}

\begin{figure}[h!] \centering 
  \includegraphics[scale=0.92]{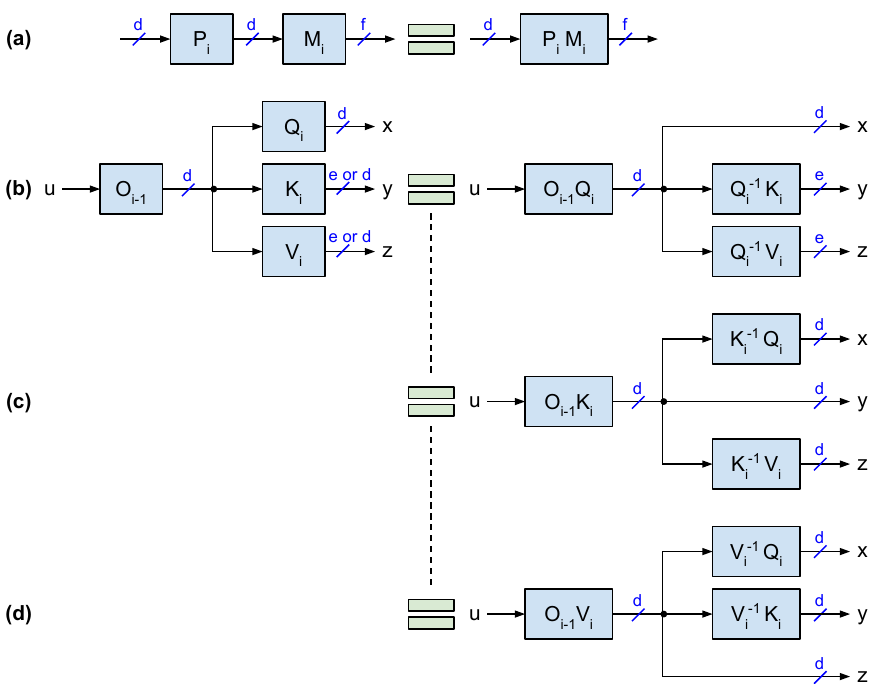}
  \caption{(a) Merging P and M; (b) eliminating Q; (c) eliminating K; (d) eliminating V.}
\label{fig2} \end{figure}

Figure \ref{fig2} details how the linear layers are merged:
\begin{itemize}[topsep=-1pt, itemsep=-1pt]
  \item Figure \ref{fig2}(a) shows how the two linear layers with weight matrices $\P$ and $\M$ are collapsed and replaced by a single linear layer with weight matrix $\M^* = \P \M$, which eliminates $d^2$ weights.
  \item Figure \ref{fig2}(b) illustrates how to merge $\Q$ into the preceding $\O$-matrix, which eliminates $d^2$ weights and requires $\Q$ to be invertible. Note that $\y = \u \O (\Q \Q^{-1}) \K = \u \O \K$ and $\z = \u \O (\Q \Q^{-1}) \V = \u \O \V$.
  \item For MHA where $e = d$, $\K$ can be removed as shown in Figure \ref{fig2}(c), which eliminates $d^2$ weights. Note that $\x = \u \O (\K \K^{-1}) \Q = \u \O \Q$ and $\z = \u \O (\K \K^{-1}) \V = \u \O \V$. This requires that $\K$ is invertible.
  \item For MHA where $e = d$, $\V$ can be removed as shown in Figure \ref{fig2}(d), which eliminates $d^2$ weights. Note that $\x = \u \O (\V \V^{-1}) \Q = \u \O \Q$ and $\y = \u \O (\V \V^{-1}) \K = \u \O \K$. This requires that $\V$ is invertible.
\end{itemize}

Table \ref{tab1} specifies how the new weight matrices ($\M^*, \Q^*, \K^*, \V^*, \O^*$) of Figure \ref{fig1} are calculated from the original ones. For the first transformer block ($i = 1$), we use the input embedding instead of $\O$ (because there is no $\O$ for $i = 1$).

\begingroup \begin{table} [h!] \centering  
\renewcommand{\arraystretch}{1.2}  
\begin{tabular}{cccc} \hline
         & Figure 1(b)     & Figure 1(c)     & Figure 1(d)    \\ \hline
  $\O^*$ & $\O \Q$         & $\O \K$         & $\O \V$        \\
  $\Q^*$ & 1 (eliminated)  & $\K^{-1} \Q$    & $\V^{-1} \Q$    \\
  $\K^*$ & $\Q^{-1} \K$    & 1 (eliminated)  & $\V^{-1} \K$    \\
  $\V^*$ & $\Q^{-1} \V$    & $\K^{-1} \V$     & 1 (eliminated) \\
  $\M^*$ & $\P \M$         & $\P \M$         & $\P \M$        \\ \hline
\end{tabular}
\caption{How to calculate the new weight matrices from the original ones for Figure \ref{fig1}.}
\label{tab1} \end{table} \endgroup

\section{Parallel transformer without skip connections}
\begin{figure} [h!] \centering  
  \includegraphics[scale=0.92]{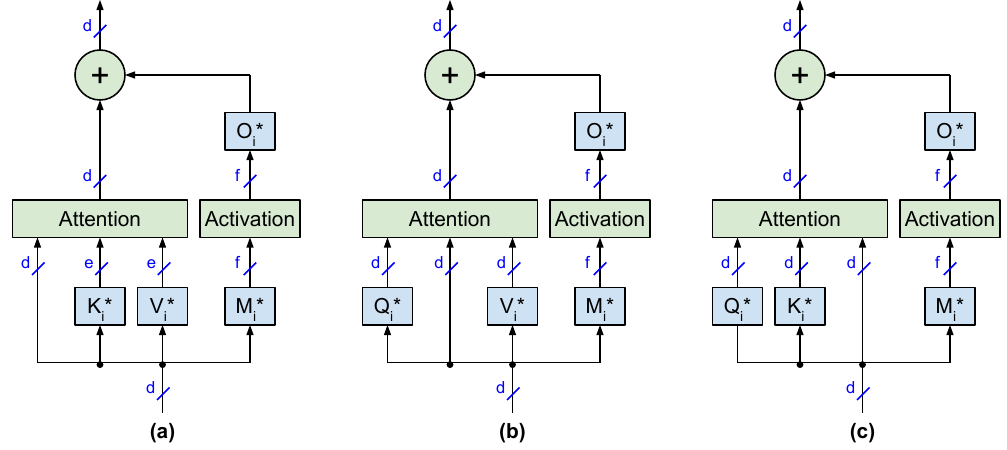}
  \caption{Parallel skipless transformers (a) without Q and P; (b) without K and P; (c) without V and P.}
\label{fig3} \end{figure}

Similar to the parallel transformer \cite{parallel}, Figure \ref{fig3} shows parallel versions of Figures \ref{fig1}(b) to (d). Here, “parallel” refers to having the attention (including its linear layers) in parallel to the FFN.

Figures \ref{fig3}(b) and (c) require that $e = d$, so they are only suitable for MHA, but not for MQA and GQA. Figure \ref{fig3}(a) is suitable for MHA, MQA, and GQA. Figure \ref{fig3}(c) is identical to the simplified transformer proposed in \cite{simplified}.

\section{Related work}
In addition to \cite{simplified}, our work is related to the lossless weight compression for back-to-back linear layers presented in \cite{matShrink}, and FlashNorm's weight fusion \cite{flashNorm}.

\section{Examples}
The table below lists the configurations and weight counts for Pythia-6.9B and Mistral-7B. For a skipless version of Mistral-7B we would save 15\% of weights after merging the Q and P linear layers into the FFN layers. For a batch 1 system that is limited by memory bandwidth, these 15\% weight savings can speed up inference by 1.17x during the autoregressive next-token-generation phase, see the table below.

\begingroup
\renewcommand{\arraystretch}{1.2} 
\begin{center} \begin{tabular}{lccl}                                                                      \hline
  \textbf{Parameter} & \textbf{Pythia-6.9B} & \textbf{Mistral-7B} & \textbf{Notes}                          \\ \hline
  Parallel attention/FFN? & parallel     & serial           & \cite{parallel}                               \\
  MHA, MQA, or GQA?       & MHA          & GQA              & \cite{vanilla, MQA, GQA}                      \\
  \verb+dim+ (aka $d$)    & \multicolumn{2}{c}{4,096}       & embedding dimension                           \\
  \verb+n_layers+         & \multicolumn{2}{c}{32}          & number of layers                              \\
  \verb+n_heads+          & \multicolumn{2}{c}{32}          & number of heads                               \\
  \verb+n_kv_heads+       & 32           & 8                & number of KV-heads                            \\
  \verb+e+ (output dim. of K, V) & 4,096 & 1,024            & \verb+e = d * n_kv_heads / n_heads+           \\
  FFN type                & MLP          & MLP with SwiGLU  & \cite{GLU}                                    \\
  FFN \verb+hidden_dim+   & 16,384       & 14,336           & FFN hidden dimension                          \\
  \verb+vocab_size+       & 50,400       & 32,000           & vocabulary size                               \\ \hline

  \multicolumn{4}{l}{\textbf{Number of weights (calculated from above parameters):}}                        \\ \hline
  Q+P weights per layer   & \multicolumn{2}{c}{33,554,432}  & \verb+2 * dim * dim+                          \\
  K+V weights per layer   & 33,554,432   &   8,388,608      & \verb+2 * dim * dim / n_heads * n_kv_heads+   \\
  FFN weights per layer   & 134,217,728  & 176,160,768      & \verb+(2 or 3) * dim * hidden_dim+            \\
  Input+output embed.     & 412,876,800  & 262,144,000      & \verb+2 * dim * vocab_size+                   \\
  \multicolumn{1}{r}{\textbf{Total weights:}} & 6.9B & 7.2B &                                               \\ \hline

  \multicolumn{4}{l}{\textbf{Weight savings and speedup after removing Q and P:}}                           \\ \hline
  Total w/o Q+P weights:                           & 5.8B           & 6.2B   & total after removing Q and P \\
  \multicolumn{1}{r}{\textbf{Weight savings:}}     & \textbf{16\%}  & \textbf{15\%}  &                      \\
  \multicolumn{1}{r}{\textbf{Possible speedup:}}   & \textbf{1.19x} & \textbf{1.17x} & assumes batch size 1 \\ \hline
\end{tabular} \end{center}
\endgroup

\section{Experiments}
Refer to \cite{tricks} for Python code that demonstrates the numerical equivalency of the weight reduction illustrated in Figures \ref{fig1}(b) and \ref{fig2}(b). The code also confirms that all square matrices of Mistral-7B are invertible.

\section{Conclusion}
A novel approach to optimizing skipless transformers by eliminating the query (Q) and post-attention projection (P) linear layers is presented. This mathematical equivalent weight fusion offers savings in computational cost, memory, and energy consumption by reducing the number of weights.

Recently published skipless transformers such as \cite{skipless2, skipless} could be retrofitted post-training with the lossless weight fusion presented here. Skipless transformers with normalization layers could first eliminate the normalization layers by fine-tuning as described in \cite{remove-norm, remove-norm2} and then apply the weight fusion described in our work.

Our work extends a recent trend in neural network design toward architectural parsimony, in which unnecessary components are removed to create more efficient models. Notable examples include RMSNorm’s simplification of LayerNorm by removing mean centering \cite{rms}, PaLM’s elimination of bias parameters \cite{PaLM}, decoder-only transformers’ omission of the encoder stack \cite{genWiki}, FlashNorm's elimination of normalization weights \cite{flashNorm}, and Slim Attention's elimination of the V-cache \cite{slimAttn}. This trend is rooted in the revolutionary shift initiated by the original Transformer model, which replaced traditional recurrence and convolutions with a more streamlined architecture \cite{vanilla}.

\begin{figure} \centering
  \includegraphics[scale=0.92]{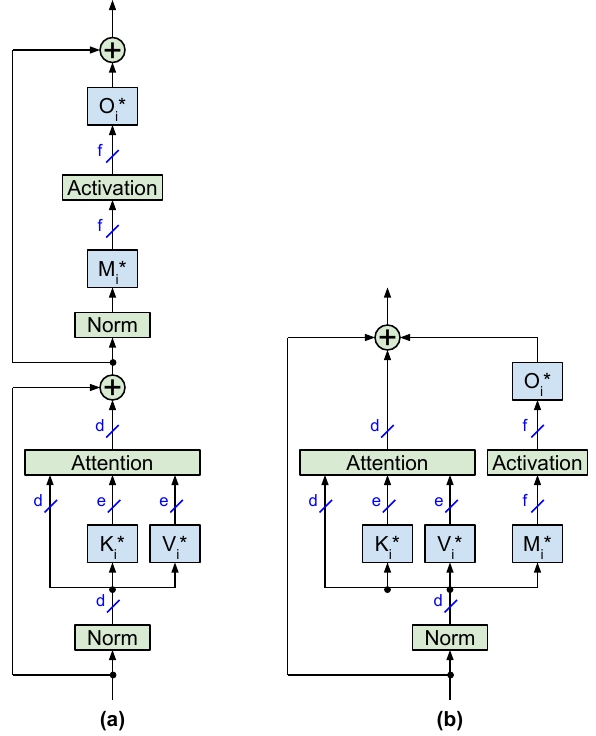}
  \caption{(a) Transformer block without Q and P; (b) version with parallel attention / FFN.}
\label{fig4} \end{figure}

Because skipless transformers are not very popular right now, future work should investigate whether removing P and Q (or K or V) is also beneficial for transformers with normalization and skip connections as illustrated in Figure \ref{fig4}. Adding normalization and skip connections again could simplify and speed up training relative to skipless transformers.

\section*{Acknowledgments}
We would like to thank \href{https://scholar.google.com/citations?user=HKft_LAAAAAJ&hl=en}{Bobby He (ETH Zürich)} and \href{https://scholar.google.com/citations?user=LlK_saMAAAAJ&hl=en}{James Martens (DeepMind)} for helpful discussions on this work.

\bibliographystyle{unsrtnat}
\bibliography{references}

\end{document}